\documentclass[conference,a4paper]{IEEEtran}
\IEEEoverridecommandlockouts

\usepackage[hidelinks]{hyperref}
\usepackage[cmex10]{amsmath}
\usepackage{amssymb,amsfonts}
\usepackage{dblfloatfix}
\usepackage{enumerate}
\usepackage[shortlabels]{enumitem}

\usepackage[ruled,vlined]{algorithm2e}
\usepackage{graphicx}
\graphicspath{{Figures/PDF/}{Figures/PNG/}}

\usepackage{booktabs}
\usepackage{siunitx}
\usepackage[numbers,compress]{natbib}
\usepackage{texnames}
\usepackage{bm,bbm}
\usepackage{orcidlink}
\usepackage{booktabs}
\usepackage[flushleft]{threeparttable}
\usepackage{subcaption}

\begin{document}

\title{\uppercase{Click2Poly: A VLM for vector mapping buildings and walls}}

\author{	
    \IEEEauthorblockN{Nicolas Girard \orcidlink{0009-0006-5630-972X}, Jawher Ben Abdallah\text{*}, Arno Gobbin\text{*}, Liuyun Duan \orcidlink{0009-0001-5632-9740}, Sacha Lepretre}
    \IEEEauthorblockA{\textit{LuxCarta}\\
    \{ngirard, jbenabdallah, agobbin, lyduan\}@luxcarta.com sacha.lepretre@gmail.com} 
    \text{*} Equal contribution
    \thanks{
        \noindent\rule{4cm}{0.4pt}
        
        This project was part of the TGI/AWS Generative AI for Geospatial Challenge and benefited from offered AWS credits. The authors wish to thank TGI and AWS and in particular Nadine Alameh from TGI (now CEO LunateAI), Phil Cooper and Dean McCormick from AWS.
    }
}

\maketitle

\begin{abstract}
    Accurate vector mapping of buildings and walls is critical for geospatial applications but remains a labor-intensive process. While recent deep learning methods have improved automatic extraction, in order to meet cartographic standards they always require a human to perform quality control and fix complex cases in the extraction. We present Click2Poly, a human-in-the-loop AI assistant designed to speed up this manual step. Extending the Florence-2~\cite{xiao2024florence} Vision Language Model (VLM), Click2Poly responds to user clicks by editing the building or wall vector layer directly. Implemented as a QGIS~\cite{QGIS_software} plugin, Click2Poly speeds up the manual editing of building and wall vector layers in a real-world production environment.
\end{abstract}


\section{Introduction}

Accurate vector maps of buildings and walls face several challenges such as a high spatial precision of the geometry, correct topology (e.g. narrow passageways between buildings must not be simplified by snapping the two polygon buildings), and simple/regularized shapes.

Fully-automatic methods to extract vector maps from satellite imagery with AI started with segmentation-based methods that first use a neural network to predict a raster mask of objects, followed by vectorization of the segmentation mask. These suffer from "blob-like" predicted masks that are ambiguous to vectorize into an accurate polygonal contour. To enforce regularity and better corner definition, the Polygonal Frame Field Learning (PFFL) method~\cite{girard2021polygonal} learns an additional output to the building segmentation: a frame field indicating the orientation of contours. The frame field allows to regularize the predicted building segmentation mask and is further used in the vectorization step. An evolution of this work increased accuracy and extended it to walls~\cite{girard2024automated}. Similarly, HiSup~\cite{xu2023hisup} exploits hierarchical supervision across masks, line segments, and vertices, improving mask reversibility so that the learned features can be polygonized into precise building footprints.

\begin{figure}[t]
    \centering
    \includegraphics[width=0.9\linewidth]{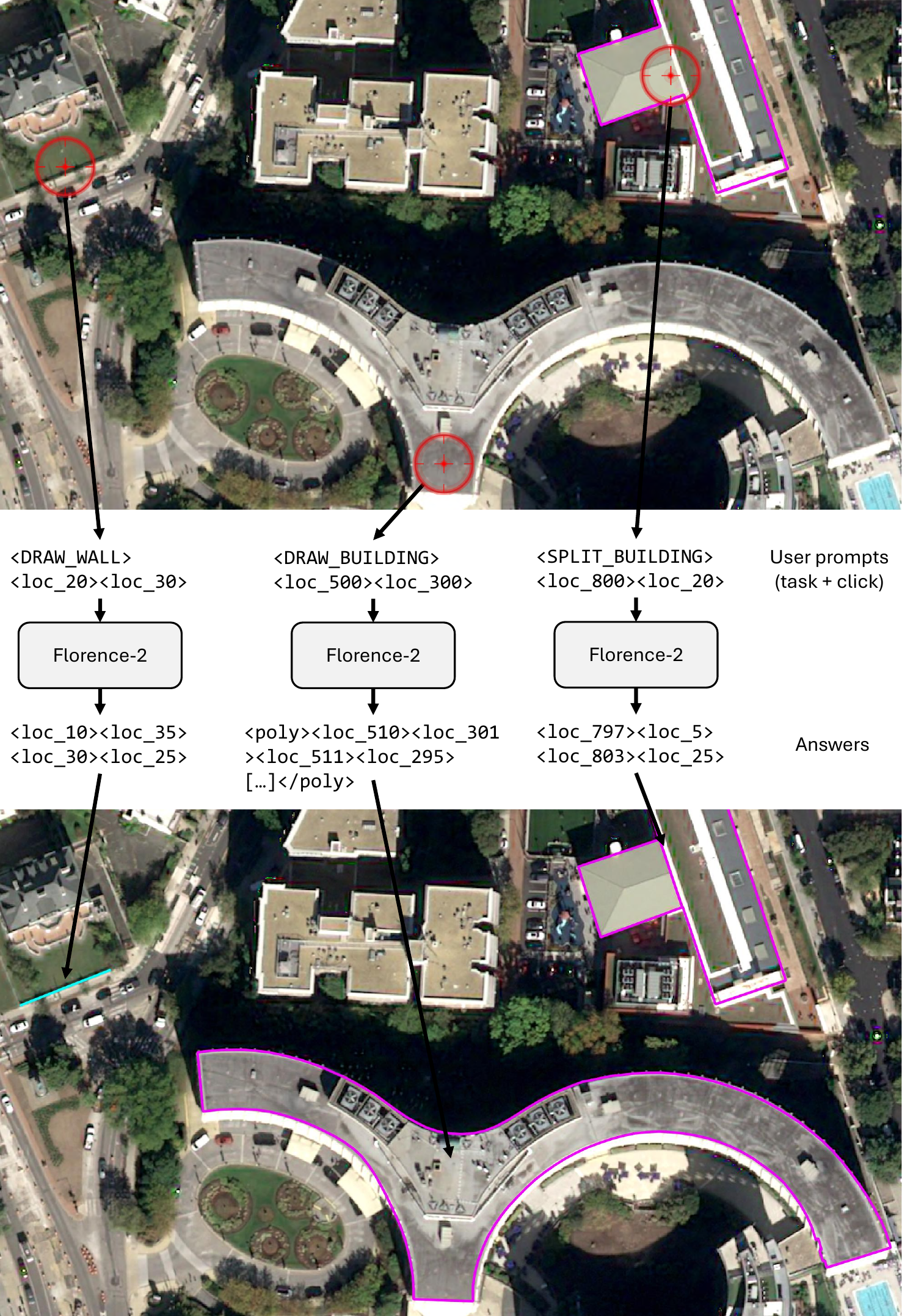}
    \caption{In the workspace the user can select a layer to edit (building or wall); select a tool among \texttt{<DRAW\_BUILDING>}, \texttt{<SPLIT\_BUILDING>}, \texttt{<DRAW\_WALL>}; and perform a click (red crosshair) to add/edit a feature. Image is a 30~cm/px Pléiades Neo in Washington D.C., USA.}
    \label{fig:image_abstract}
\end{figure}

Another branch of work propose geometric deep learning architectures that directly predict the vector map. Polygon-RNN~\cite{Castrejon2017PolygonRNN} and Polygon-RNN++~\cite{acuna2018efficient} introduced a CNN-RNN model that sequentially predicts polygon vertices in a user-defined bounding box. This approach allowed user interactivity by updating the predicted sequence if the user choose to move a vertex. Re:PolyWorld~\cite{zorzi2023re} proposed a Graph Neural Network with optimal transport to enforce closed loops and with multiple vertex instancing to allow several polygons to reuse the same vertex. Recent transformer-based methods like Pix2Poly~\cite{adimoolam2025pix2poly} and RoIPoly~\cite{zhang2025roipoly} further refine this by treating vectorization as token sequence generation and query-based set prediction respectively.

In parallel to fully-automatic methods, interactive segmentation models were developed. The Segment Anything Model (SAM)~\cite{kirillov2023segment} introduced general promptable raster segmentation. Several works adapted SAM to extracting buildings. SAMPolyBuild~\cite{wang2024sampolybuild} predicts extra raster maps to help a post-process vectorization. PolyFootNet~\cite{PolyFootNet} focuses on segmenting simultaneously roofs and footprints in off-nadir imagery. It predicts polygons by direct coordinate computing and a mask-guided connecting strategy similar to HiSup~\cite{xu2023hisup}. The GeoAI plugin~\cite{geoai_qgis_plugin} is a more production-ready tool adapting SAM.

The most important aspect of mapping buildings (and walls) is the accuracy of the final vertex placements. The most simple end-to-end method will have an edge over more complex ones involving extra outputs and post-processing steps. Additionally, shapes of buildings look regular at first glance (e.g. would be well reconstructed as a union of rectangles) but actually a non-negligible amount of buildings have non-regular shapes and thus a general method not introducing any kind of fixed bias of the final shape is preferred. For these reasons our work thus leverages the Florence-2~\cite{xiao2024florence} Vision Language Model as it can predict location tokens directly and is supervised by a single cross-entropy loss, making the accurate placement of vertices its only objective. It has no fixed bias on the final shapes it can predict while allowing to learn shape bias from the data which is useful to predict shapes under occlusion. The Seq2Seq nature of Florence-2 makes it very adaptable and allowed us to design an interactive system to help users create or fix maps of buildings and walls faster using clicks. In addition, this interactive system is a great stepping-stone towards a future fully-automatic mode that still allows user-in-the-loop which is compulsory in a real-world production environment.

\noindent Our contributions are:
\begin{enumerate}[(i),itemsep=-1pt,topsep=1pt]
    \item a model predicting polygon buildings on satellite imagery
    \item a system able to split building blocks into individual buildings
    \item a model predicting linear features for walls on satellite imagery
    \item with human-in-the-loop for better control
\end{enumerate}

\section{Method}

Our work extends the Florence-2~\cite{xiao2024florence} model by training it on three new tasks (tools): \texttt{<DRAW\_BUILDING>}, \texttt{<SPLIT\_BUILDING>}, \texttt{<DRAW\_WALL>}. The prompt is composed of the task plus the user click in the image where the user wants to extract a building/wall (akin to SAM~\cite{kirillov2023segment}) or split a building. See Figure~\ref{fig:image_abstract} for an overview of the method.

\subsection{Dataset}

The building dataset contains 377,207 building polygons across 202 geographic areas (159 for training and 43 for validation), covering a total extent of approximately 1,344~km\textsuperscript{2}. As for the walls dataset, it contains 197,234 wall linestrings across 117 geographic areas (94 for training and 23 for validation), covering a total extent of approximately 201 km². Both dataset use 30~cm/px satellite imagery. The areas are patched into smaller images of $768\times768$~px to match Florence-2 input size. We used standard image augmentation techniques such as rotation, rescaling and color perturbations.

For training, the raw data (image + annotations) needs to be transformed to simulate a user-model interaction. For \texttt{<DRAW\_BUILDING>} a random building contour is chosen in the image and a random point is sampled within that building contour to act as a user input click. A building contour can either be an exterior contour (shell) or an interior contour (hole) to allow drawing buildings with holes using multiple successive click interactions. For \texttt{<SPLIT\_BUILDING>} a random edge (two vertices connected together in a building polygon's contour) is chosen and a random point is sampled around that edge to act as a user input click. Similarly for \texttt{<DRAW\_WALL>} a random edge (two vertices connected together in a wall linestring) is chosen and a random point is sampled around that edge to act as a user input click.

All coordinates (user clicks and ground truth geometries) are translated into Florence-2's vocabulary using its location tokens. The training sample is thus composed of an $768\times768$~px image, a prompt (task + simulated user click tokenized coordinates), and an answer (polygon tokenized coordinate sequence or edge tokenized coordinates depending on the task). 

\subsection{Training}

To avoid training the full Florence-2 model of 230~M parameters we used the LoRA method~\cite{hu2022lora} with a rank 256 on major layers, lowering the number of trainable parameters to 70~M. We trained one adapter per task. \texttt{<DRAW\_BUILDING>} on 8 A100 GPUs with a batch size of 8 for 7 days. \texttt{<SPLIT\_BUILDING>} on 8 A100 GPUs with a batch size of 7 per GPU for 27~days. \texttt{<DRAW\_WALL>} for walls on a single L40S GPU with a batch size of 10 for 14~days.

\subsection{Implementation details}

The system is implemented in a server-client fashion with multiple clients able to connect to the same server. For inference on large satellite images, Click2Poly works on a grid of overlapping $768\times768$~px cells. For each user click the cell whose center is closest to the click is selected. The image is cropped and fed to the image encoder of Florence-2 and then the Seq2Seq part of Florence predicts vertices in sequence. The intermediate image encodings of the cell are cached so that the next user click choosing that cell can skip the image encoder and produce an output faster. In terms of performance, an image encoding request round-trip takes about 1.3~s and a polygon prediction request round-trip takes about 0.8~s (the server running on a RTX 3060 GPU).

\subsection{Experimental setup}

The main objective of Click2Poly is to speed up the compulsory manual correction of automatic extractions in a real-world production setting. In order to measure this speedup we first run our latest automatic building extraction model (an evolution of \cite{10642761}) on 4 new test areas (never seen by our automatic extraction model nor Click2Poly). We then measure the time it takes an expert operator to fix all the issues for two scenarios: the first one without Click2Poly (using standard QGIS tools + internal non-AI digit helper tools) and the second one adding Click2Poly to the available tools. We have to account for a familiarity bias: if an operator first fixes an area without Click2Poly and then does the same exercise using Click2Poly, part of the speedup could come from being already familiar with the area and knowing which geometries to fix and their correct shape. To account for this, we used the Latin square design for the experiment: 4 operators each fixed the 4 test areas, alternating which method is used per area.

During the experiment, each operator's results (standard and Click2Poly) are checked by an annotation manager to make sure all the resulting vector maps pass quality control. In addition to measuring operator times, we also measured operator agreement using the IoU (Intersection over Union) between each pair of operator result per area.

The 4 test areas were chosen to be diverse and cover various types of urban landscapes:
\begin{itemize}
    \item A: 0.5~km² in Luanda, Angola of a 30~cm/px Pléiades Neo image ($\approx700$ buildings, sparse residential).
    \item B: 0.25~km² São Paulo, Brazil of a 30~cm/px Pléiades Neo image ($\approx400$ buildings, city center).
    \item C: 0.5~km² Rosario, Philippines of a 30~cm/px Pléiades Neo image ($\approx3,000$ buildings, dense residential).
    \item D: 2.5~km² Kampala, Uganda of a 50~cm/px Pléiades image ($\approx1,000$ buildings, commercial).
\end{itemize}

\section{Results and Discussion}

\begin{figure}[h]
    \begin{subfigure}[b]{0.49\linewidth}
        \includegraphics[width=\linewidth]{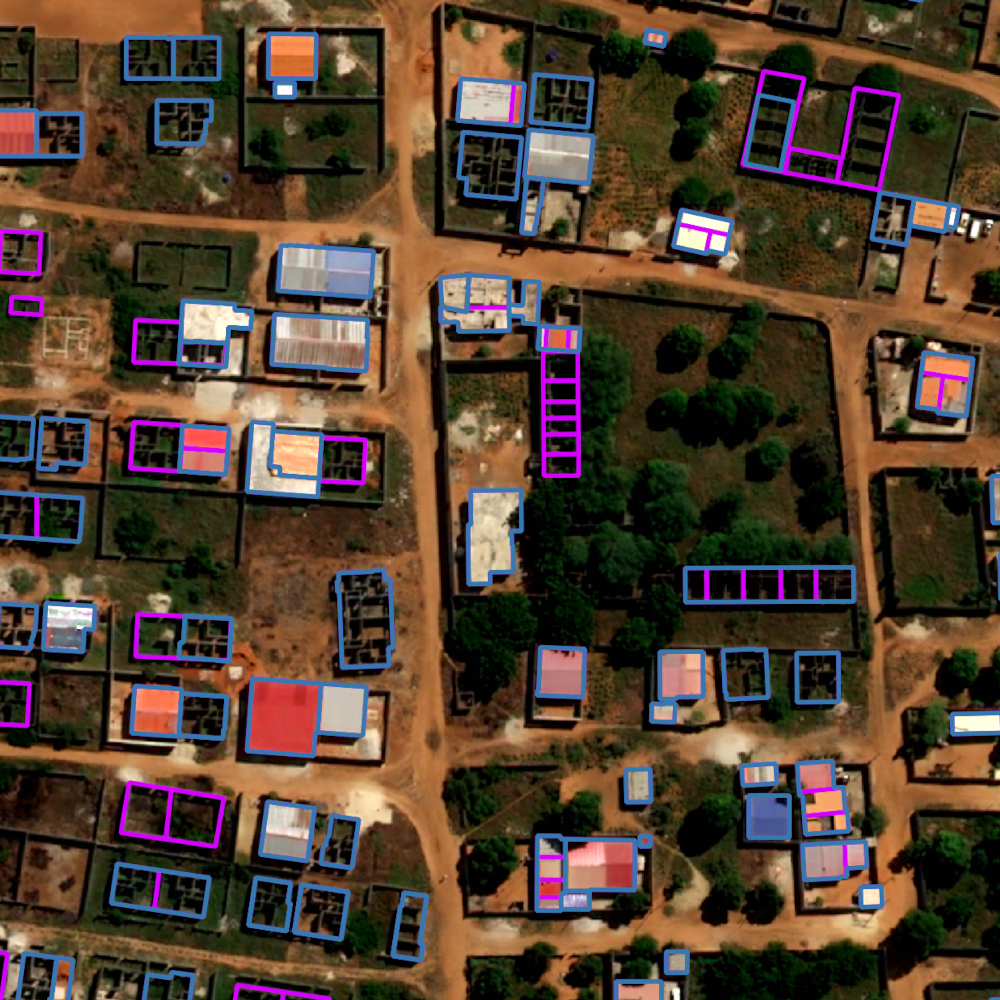}
        \caption{Luanda, Angola}
        \label{fig:main_experiment_a}
    \end{subfigure}
    \hfill
    \begin{subfigure}[b]{0.49\linewidth}
        \includegraphics[width=\linewidth]{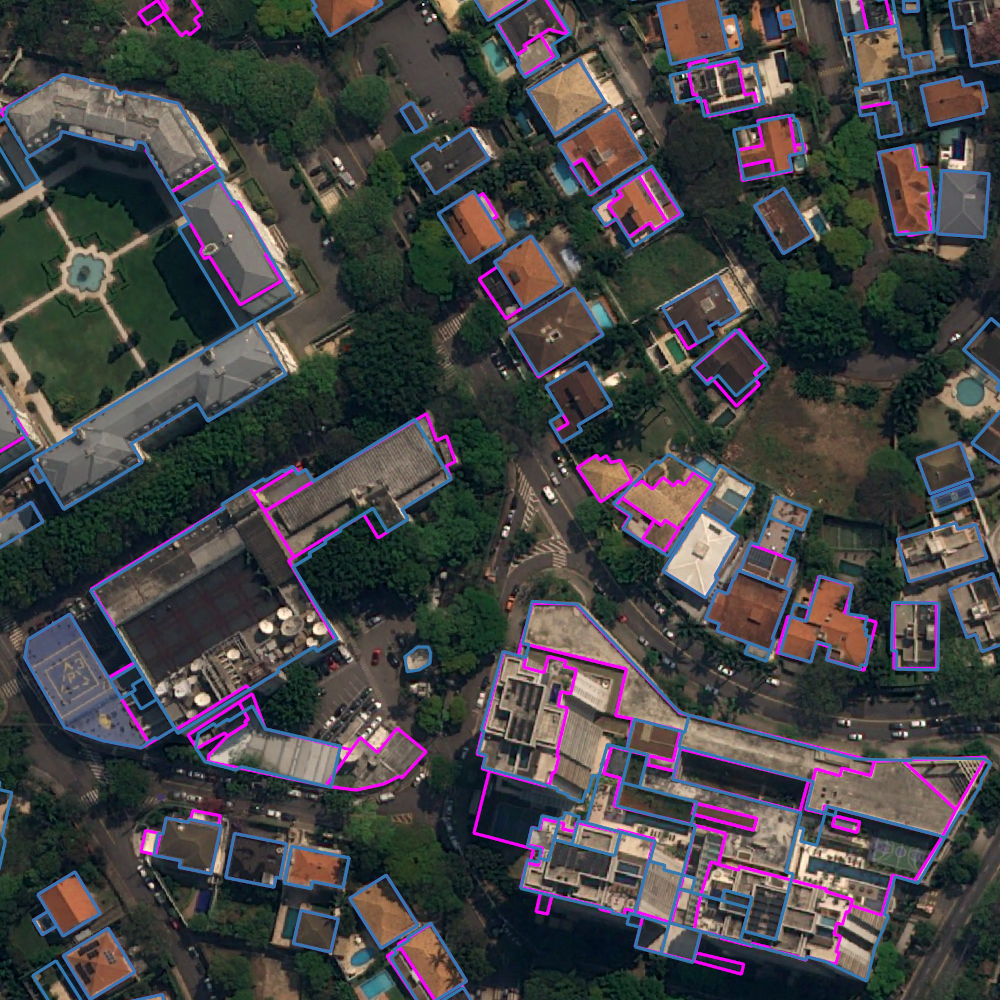}
        \caption{São Paulo, Brazil}
        \label{fig:main_experiment_b}
    \end{subfigure}
    
    \begin{subfigure}[b]{0.49\linewidth}
        \includegraphics[width=\linewidth]{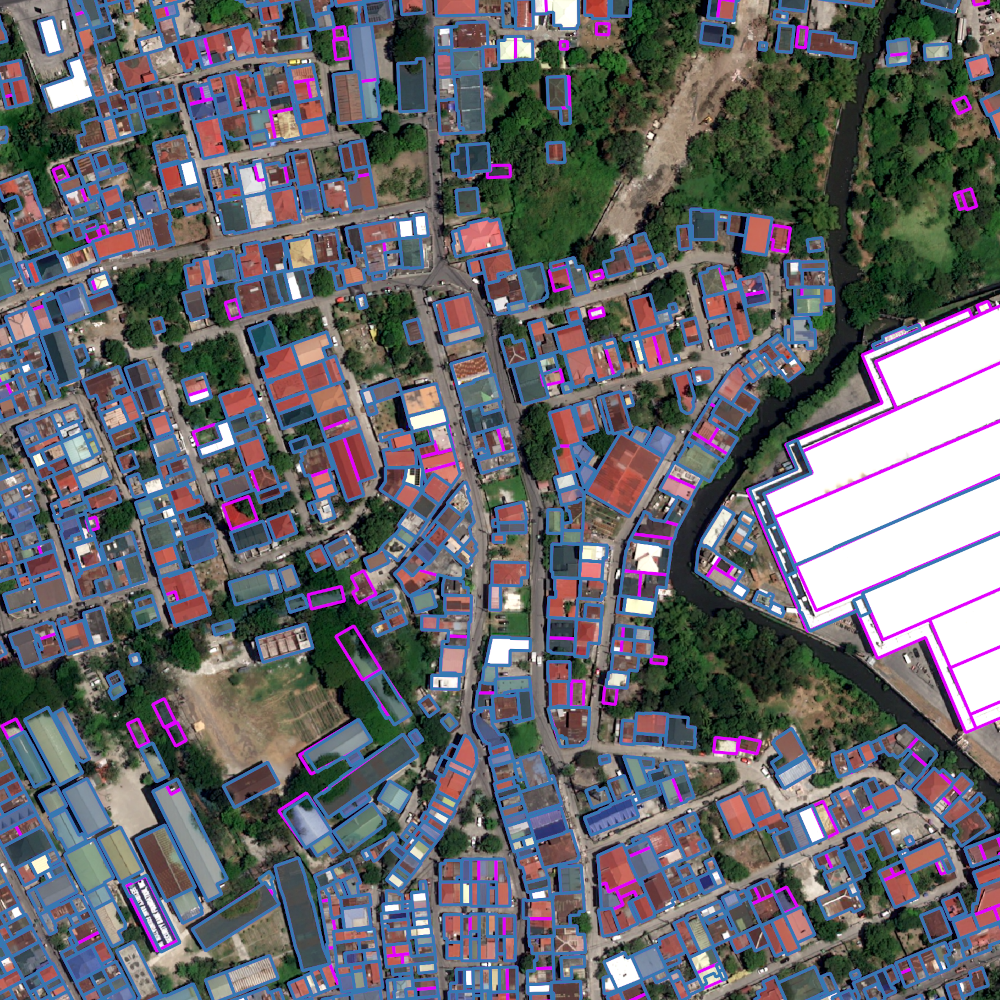}
        \caption{Rosario, Philippines}
        \label{fig:main_experiment_c}
    \end{subfigure}
    \hfill
    \begin{subfigure}[b]{0.49\linewidth}
        \includegraphics[width=\linewidth]{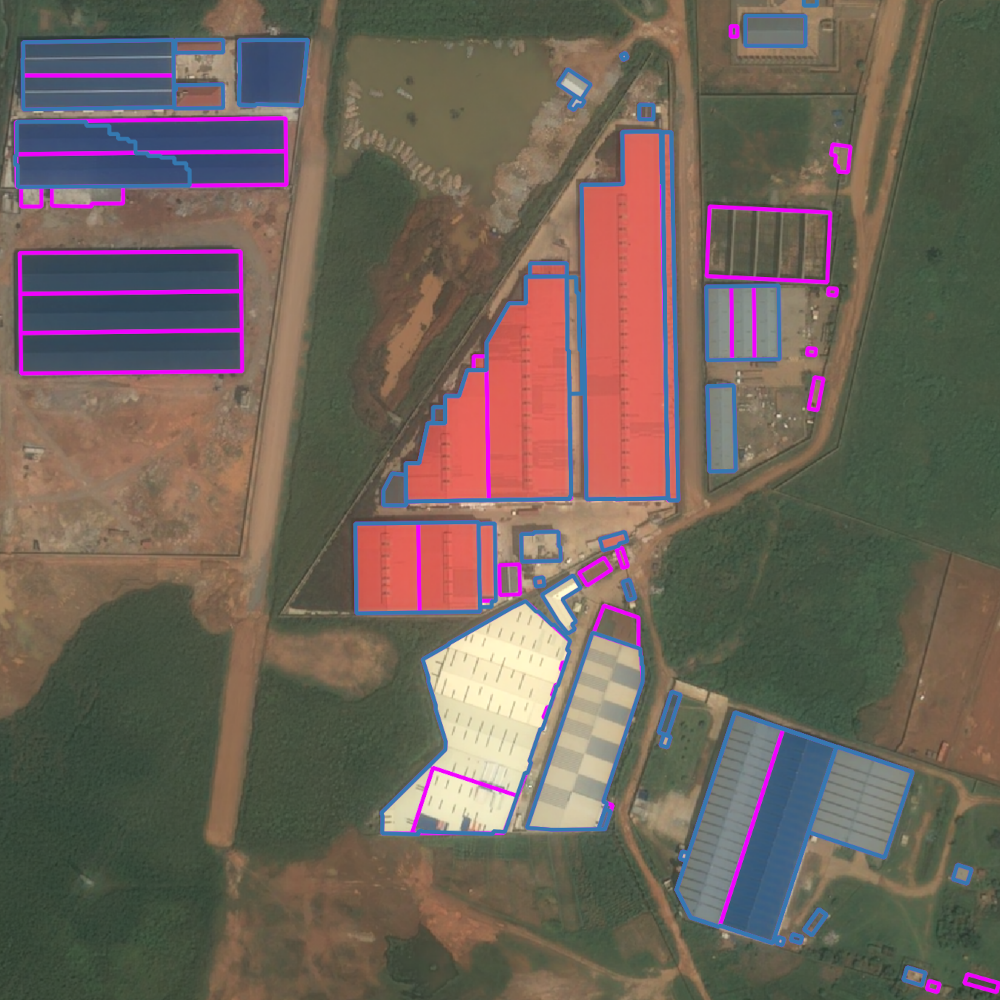}
        \caption{Kampala, Uganda}
        \label{fig:main_experiment_d}
    \end{subfigure}
    
    \caption{Crops of test images. In blue we show the automatic extractions. In purple are the fixes made with Click2Poly.}
    \label{fig:main_experiment}
\end{figure}

\textbf{Experimental results.} Visual results on the test areas are shown in Figure~\ref{fig:main_experiment}, showing where Click2Poly was used to fix the automatic extraction and the various cases it can handle (including buildings in construction). Figure~\ref{fig:image_abstract} also shows some predictions working on a curved building. Timing results are reported in Table~\ref{tab:latin_square_results}. Speedup differences are due to the variable size, number of buildings, and difficulty of the areas. Generally, Click2Poly increases the speed of manual correction with a total speed-up of $\times1.53$. However, area B in a city center with multi-level complex buildings on an off-nadir image was slower when using Click2Poly because some of its results had to be manually fixed by standard tools. Area C had the best speed-up, showing that the model works very well in a dense built-up area as long as the buildings are not multi-level and the image is closer to nadir. 

\begin{table}[htbp]
\centering
\caption{Time (in minutes) for manual correction of automatic building extractions across operators and test areas using standard tools versus Click2Poly.}
\label{tab:latin_square_results}
\begin{tabular}{lccccc}
\toprule
\textbf{Operator} & \textbf{Area A} & \textbf{Area B} & \textbf{Area C} & \textbf{Area D} & \textbf{Total} \\
\midrule
Op1 & 40 (S) & 69 (C) & 125 (S) & 56 (C) & \\
Op2 & 49 (S) & 60 (C) & 116 (S) & 58 (C) & \\
Op3 & 25 (C) & 58 (S) & 48 (C) & 81 (S) & \\
Op4 & 25 (C) & 62 (S) & 57 (C) & 76 (S) & \\
\midrule
\textbf{Mean (S)} & 44.5 & 60 & 120.5 & 78.5 & 303.5 \\
\textbf{Mean (C)} & 25 & 64.5 & 52.5 & 57 & 199 \\
\midrule
\textbf{Speedup} & $\times1.78$ & $\times0.93$ & $\times2.30$ & $\times1.38$ & $\times1.53$ \\
\bottomrule
\end{tabular}
\begin{tablenotes}
    \small
    \item S = Standard tools; C = Click2Poly. Speedup calculated as $\frac{\overline{\text{Standard}}}{\overline{\text{Click2Poly}}}$.
\end{tablenotes}
\end{table}

\textbf{Operator agreement}. We show agreement between operators for each area in Table~\ref{tab:iou_results}. Firstly, these results show that mapping is inherently ambiguous as 4 different operators produce 4 slightly different maps, making it difficult to define a true ground truth for deep learning training. Secondly, the results generally have above 90\% IoU agreement, making sure the various timings in Table~\ref{tab:latin_square_results} are comparable. Lastly, we observe that area B has the least amount of agreement with $89.1\%$ IoU, likely because it is a more difficult area.

\begin{table}[htbp]
\centering
\caption{Intersection over Union (IoU) between operator results for each test area. Only the upper triangle is shown due to symmetry. Average IoU agreement per area: $A=90.7\%$, $B=89.1\%$, $C=96.5\%$, $D=95.9\%$.}
\label{tab:iou_results}
\resizebox{\linewidth}{!}{%
\begin{tabular}{rccc|rccc}
\toprule
\textbf{Area A} & \textbf{Op2 (S)} & \textbf{Op3 (C)} & \textbf{Op4 (C)} & \textbf{Area B} & \textbf{Op2 (C)} & \textbf{Op3 (S)} & \textbf{Op4 (S)} \\
\midrule
Op1 (S) & 91.59\% & 91.22\% & 89.14\% & Op1 (C) & 90.25\% & 89.22\% & 87.73\% \\
Op2 (S) &         & 90.49\% & 88.69\% & Op2 (C) &         & 89.45\% & 87.74\% \\
Op3 (C) &         &         & 93.16\% & Op3 (S) &         &         & 90.02\% \\
\midrule
\textbf{Area C} & \textbf{Op2 (S)} & \textbf{Op3 (C)} & \textbf{Op4 (C)} & \textbf{Area D} & \textbf{Op2 (C)} & \textbf{Op3 (S)} & \textbf{Op4 (S)} \\
\midrule
Op1 (S) & 96.26\% & 96.58\% & 95.33\% & Op1 (C) & 96.80\% & 96.34\% & 95.81\% \\
Op2 (S) &         & 96.62\% & 96.97\% & Op2 (C) &         & 96.38\% & 95.37\% \\
Op3 (C) &         &         & 97.18\% & Op3 (S) &         &         & 94.64\% \\
\bottomrule
\end{tabular}%
}
\end{table}

\begin{figure}[h]
    \centering
    \includegraphics[width=\linewidth]{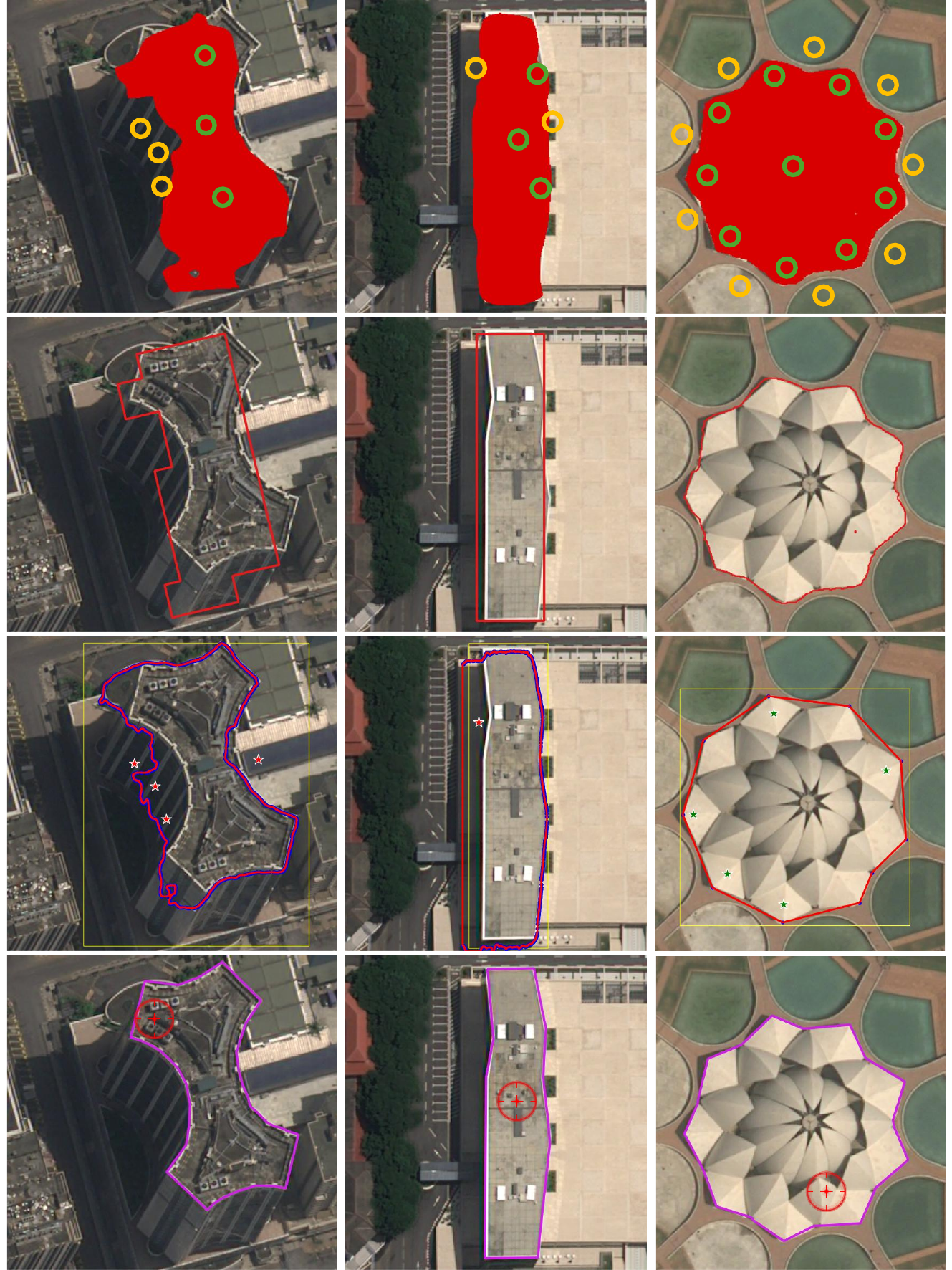}
    \caption{Results from different interactive methods. First row: GeoAI plugin~\cite{geoai_qgis_plugin} running the SAM~3 model~\cite{carion2025sam3segmentconcepts}. Second row: vectorization by the GeoAI plugin~\cite{geoai_qgis_plugin}. Third row: SAMPolyBuild~\cite{wang2024sampolybuild}. Fourth row: \textbf{Click2Poly}. Images are 30~cm/px Pléiades Neo image (first two columns in Abidjan, Ivory Coast and third column in New Delhi, India)}
    \label{fig:visual_comparisons}
\end{figure}

\textbf{Comparisons.} To showcase the advantage of predicting geometry directly by deep learning, in Figure~\ref{fig:visual_comparisons} we compare with the GeoAI plugin~\cite{geoai_qgis_plugin} running the SAM~3 model~\cite{carion2025sam3segmentconcepts} and with SAMPolyBuild~\cite{wang2024sampolybuild}. SAM allows for a bounding box and multiple foreground or background clicks to constrain the segmentation. We used this feature to get the best results in both GeoAI plugin~\cite{geoai_qgis_plugin} and SAMPolyBuild~\cite{wang2024sampolybuild}. We also tried the optional regularized vectorization of the GeoAI plugin~\cite{geoai_qgis_plugin} in the first two columns. We observe that Click2Poly is able to follow contours more accurately, using a single click, while other methods could not do the same even when using multiple clicks for more guidance. 

\begin{figure}[h]
    \centering
    \includegraphics[width=0.9\linewidth]{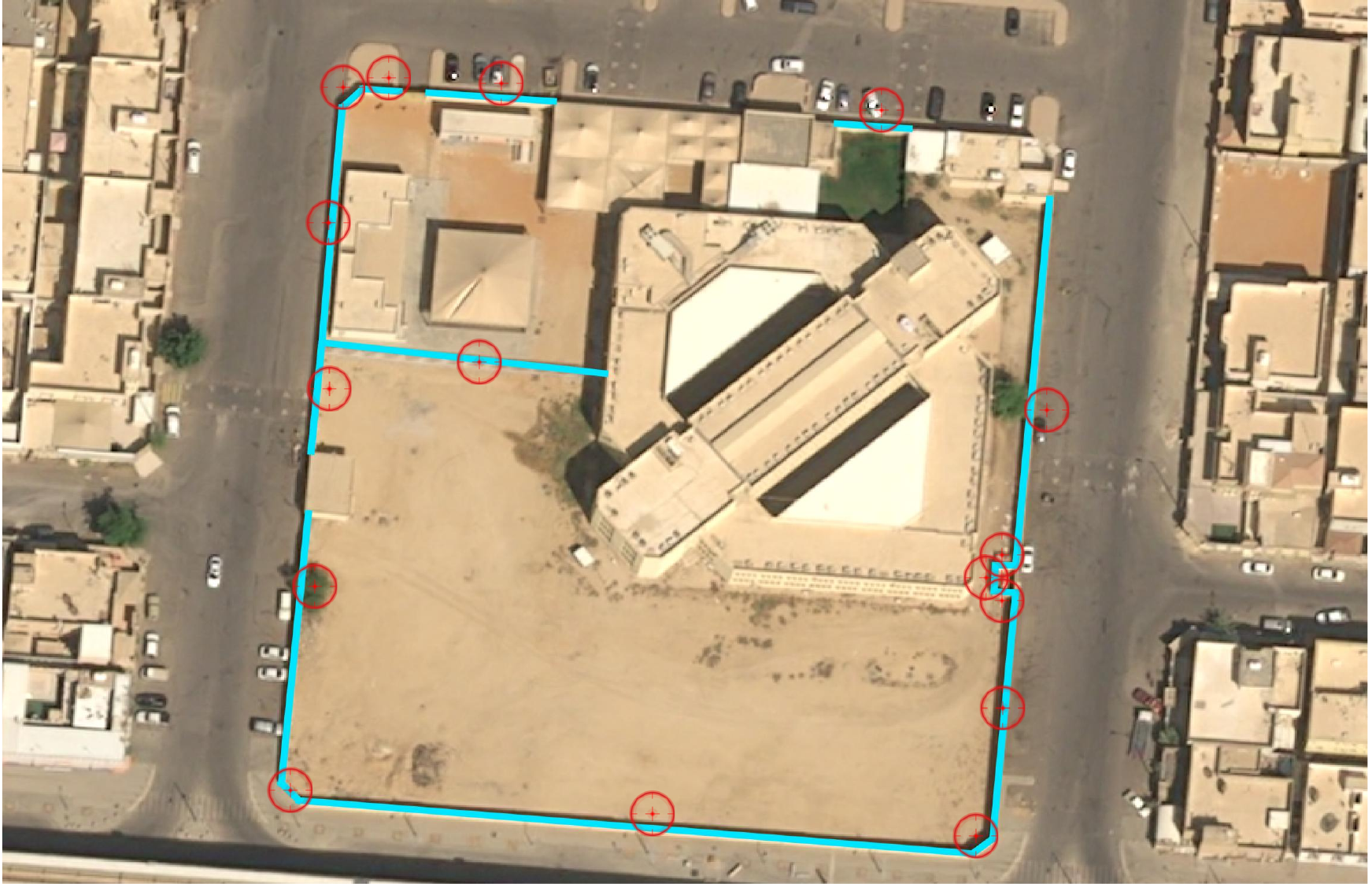}
    \caption{Wall extraction, with the click per wall section shown on a 30~cm/px Pléiades Neo image in Riyad, Saudi Arabia.}
    \label{fig:wall_showcase}
\end{figure}

\textbf{Walls}. As walls are mapped using linear features, the GeoAI plugin~\cite{geoai_qgis_plugin} and SAMPolyBuild~\cite{wang2024sampolybuild} being segmentation-based cannot output anything meaningful. We thus just show Click2Poly's results for walls in Figure~\ref{fig:wall_showcase}, predicting accurate walls directly.

\begin{figure}[h]
    \centering
    \includegraphics[width=0.9\linewidth]{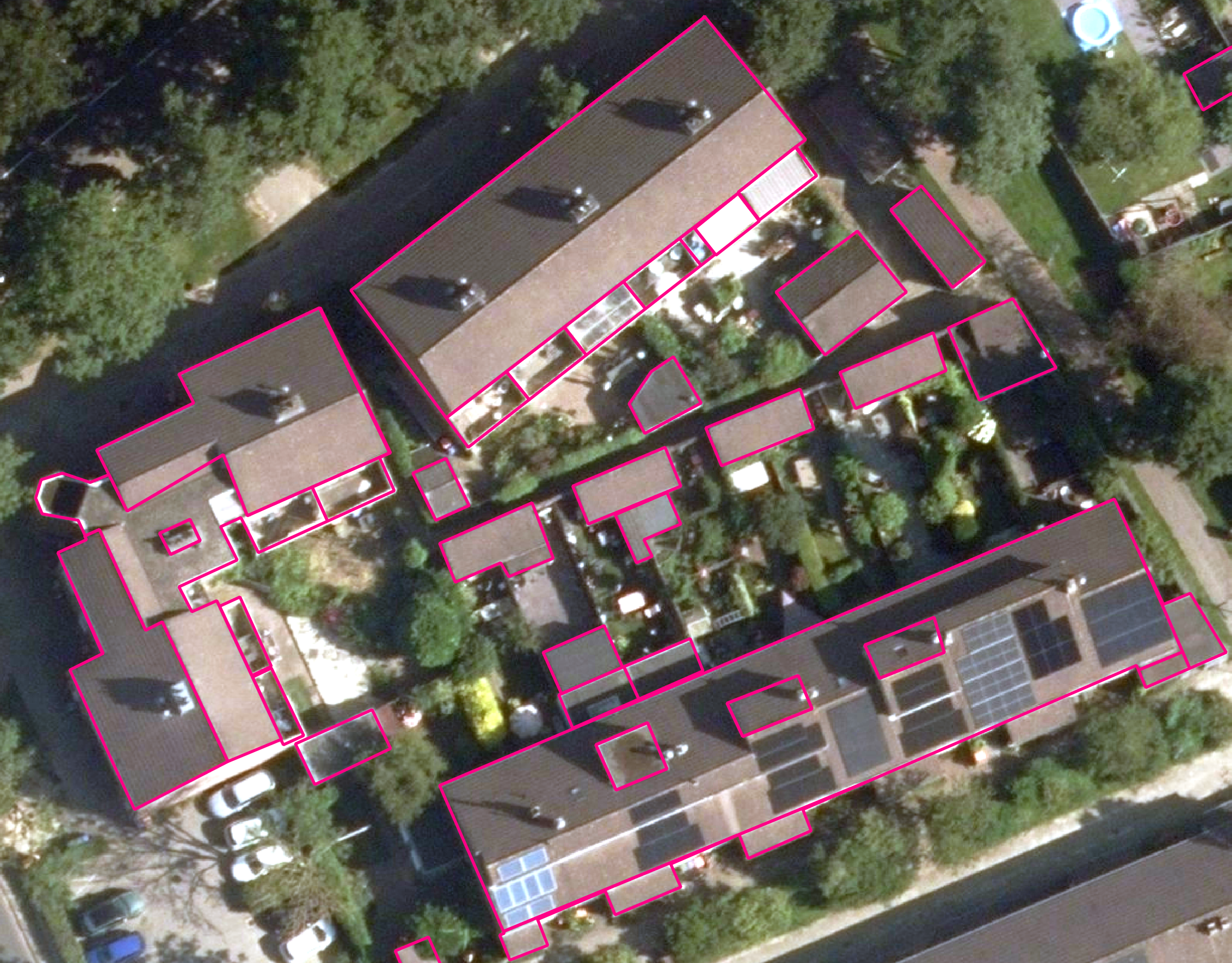}
    \caption{Click2Poly results on an aerial image of 7.5~cm/px in Amersfoort, Netherlands.}
    \label{fig:aerial_showcase}
\end{figure}

\textbf{Aerial.} We also tried Click2Poly on aerial imagery of 7.5~cm/px, a much higher resolution the model was trained on. An example can be seen in Figure~\ref{fig:aerial_showcase}. The model generalized well to higher resolution, showing it generally learns to follow contours of buildings, regardless of resolution. As the model is free to predict any position for all vertices, we can see it can capture non-rectangular buildings well, does not over-regularize, still predicts the lowest number of vertices possible and captures fine details.

\textbf{Limitations.} Florence-2's input size being $768\times768$~px, only the visible part of large buildings extending beyond this size is extracted. The deep learning model also does not snap predicted geometry and thus can result in overlapping geometries. 

\section{Conclusion}

We propose Click2Poly, a simple interactive Vision Language Model extending Florence-2~\cite{xiao2024florence} and implemented as a QGIS plugin that directly predicts building and wall geometries from overhead imagery with no post-processing. This single step approach avoids error accumulation and produces precise and sparse contours.

In future work we will aim to solve the snapping of predicted geometries to their surrounding, and the issue of large buildings. We will also extend the capabilities of Click2Poly towards a fully-automated mode.

\small
\bibliographystyle{IEEEtranN}
\bibliography{references}

\end{document}